\documentclass{article}

\usepackage{graphics}
\usepackage[linesnumbered,algoruled]{algorithm2e}
\usepackage{amsmath}
\usepackage{amssymb}

\begin{document}

\title{Assumption-Based Planning}

\author{Damien Pellier, Humbert Fiorino\\
Laboratoire Leibniz (CNRS - INPG - IMAG) \\
46, avenue F\'elix Viallet, F-38031, Grenoble\\
Email: \{Damien.Pellier, Humbert.Fiorino\}@imag.fr
}

\date{November 2004}
\maketitle

\begin{abstract}
The purpose of the paper is to introduce a new approach of planning called Assumption-Based Planning. This approach is a very interesting way to devise a planner based on a multi-agent system in which the production of a global shared plan is obtained by conjecture/refutation cycles. Contrary to classical approaches, our contribution relies on the agents reasoning that leads to the production of a plan from planning domains. To take into account complex environments and the partial agents knowledge, we propose to consider the planning problem as a defeasible reasoning where the agents exchange proposals and counter-proposals and are able to reason about uncertainty. The argumentation dialogue between agents must not be viewed as a negotiation process but as an investigation process in order to build a plan. In this paper, we focus on the mechanisms that allow an agent to produce ``reasonable'' proposals according to its knowledge.
\end{abstract}

\section{Introduction}

This paper tackles the problem of devising an intelligent agent able to elaborate plans under partial knowledge and/or to produce plans that partially contradict its knowledge. In other words, in order to reach a goal, such an agent is able to provide a plan {\it which could be executed if certain conditions were met}. Unlike ``classical'' planners, the planning process does not fail if some conditions are not asserted in the knowledge base, but rather proposes an Assumption-Based Plan or {\it conjecture}. Obviously, this conjecture must be {\it reasonable}: the goal cannot be considered ``achieved'' and the assumptions must be as few as possible because they become new goals for the other agents. For instance, suppose that a door is locked: if the agent seeks to get into the room behind the door and the key is not in the lock, the planning procedure fails even though the agent is able to fulfill 100\% of its objectives behind the door. Another possibility is to suppose for the moment that the key is available and then plan to open the door whereas finding the key might become a new goal to be delegated. To that end, we designed a planner that relaxes some restrictions regarding the applicability of planning operators.\\

The Assumption-Based Planning process is based on the concept of proof validity that can be considered as an exchange of proposals and counter-proposals. According to \cite{lakatos-76}, a correct proof does not exist in the absolute. At any time, an experimentation or a test can refute a proof. If one single test leads to a refutation, the proof is reviewed and it is considered as mere conjecture which must be repaired in order to reject this refutation and consequently becomes less questionable. The new proof can be subsequently tested and refuted anew. Therefore, the proof elaboration is an iterative non monotonous process of conjectures/refutations/repairs. \\

The same goes for our approach: each agent can refine, refute or repair the current conjecture. If the reparation of a previously refuted plan succeeds, it becomes more robust but it can still be refuted later. If the reparation of the refuted plan fails, agents leave this part of the reasoning and explore another conjecture: ``bad'' conjectures are ruled out because there is no agent able to push the process further. Finally, as in an argumentation with opponents and proponents, the current conjecture is considered as an acceptable solution when the proposal/counter-proposal cycles end and all assumptions have been removed. \\

The conjecture \--- refutation cycles can be illustrated by the following informal dialog:
\begin{itemize}
\item[$Ag_{1}$:] ``If I had fuel, I could load the passenger {\it fred} at {\it downtown}, move the taxi {\it t} from {\it downtown} to {\it park} and unload it at {\it park}, but I have no fuel'': (1) initial refinement of the goal: the lack of fuel does not lead to the planning failure but becomes an assumption to be removed;
\item[$Ag_{2}$:] ``I can provide you fuel'': (2) refinement of the conjecture: "has fuel" is no more an assumption.
\item[$Ag_{1}$:] ``Thank you!'' (3)
\item[$Ag_{2}$:] ``But you need to pay the taxi to move the passenger $fred$ from {\it downtown} to {\it park}'': (4) refutation of the conjecture;
\item[$Ag_{2}$:] ``Therefore, you can load the passenger {\it fred} at $donwtown$, pay the taxi {\it t} and move it from {\it downtown}  to {\it park}'': (5) repairing of the conjecture by adding actions to execute;
\item[$Ag_{1}$:] ``Yes, if I had money... But sorry, I cannot pay'': (6) refutation of the conjecture;
\item[$Ag_{3}$:] ``OK, I'll pay the taxi for you'': (7) refinement of the conjecture, "has money" is no more an assumption. \\
\end{itemize}

This informal example shows how agents iteratively refine (1, 2, 7), refute (4, 6) or repair (5) the current conjecture in order to produce  an acceptable plan: $Ag_{1}$ loads {\it fred} in $t$ at {\it donwtown}, $Ag_{2}$ refuels $t$, $Ag_{3}$ pays the taxi, $Ag_{1}$ moves $t$ from {\it downtown} to {\it park}, and $Ag_{1}$ unloads {\it fred} at {\it park}. As a matter of fact, Assumption-Based Planning raises many challenging issues: how to plan with incomplete information? Which reasonable assumptions can be put forward in order to reach a given goal? How to define the conjecture/refutation protocol so as to converge to an acceptable solution?\\

In this paper, we focus on the Assumption-Based Planning algorithm, i.e. on how one agent elaborates a conjecture. In section \ref{cp-versus-abp}, we briefly summarize the classical planning approach to introduce our Assumption-Based Planning model. In sections \ref{Algorithm} and \ref{planning-algo}, we describe our own planning algorithm. Then, we discuss (section \ref{discussion}) the properties of our approach. The last section is dedicated to related works (section \ref{related-works}).

\section{Classical planning versus Assumption-Based Planning}
\label{cp-versus-abp}

\subsection{Classical planning model overview}

Classical planning can be defined by a tuple $\langle {\cal{G}}, {\cal{E}}, {\cal{A}} \rangle$: $\cal{G}$, is a goal description (i.e., a set of world states), $\cal{E}$ is a partial description of the world (i.e., the agent's knowledge) and $\cal{A}$ is a description of the actions that an agent can execute. $\cal{E}$ and $\cal{G}$ are described in propositional logic. For instance, the description of the world state can be written as follows:
\begin{eqnarray*}
& \{ \ at(cab38, downtown), at(fred, downtown), &\\
& hasfuel(cab38, 10) \ \} &
\end{eqnarray*}

The goal of the agent is described by a set of knowledge defining the world  state to be reached after a plan execution. In our example, the goal is reduced to a set containing only one proposition:
\begin{equation*}
\{ \ at(fred, park) \ \}
\end{equation*}
In general, an action is described by an operator defined by:
\begin{itemize}
\item a {\it name}, with parameters;
\item a {\it precondition list} (i.e, the knowledge that must hold to apply the action);
\item a {\it del list} (i.e., the knowledge that does not hold after the action execution);
\item an {\it add list} (i.e., the knowledge that holds after the action execution).\\
\end{itemize}

For example, consider a taxi {\it cab38} at {\it downtown} and a passenger {\it fred} at {\it downtown} too. The goal submitted to the team is to move {\it fred} from his initial location to {\it park}. Considering the following actions:\begin{itemize}
\item load a passenger {\it p} in a taxi {\it t} at a specific location {\it x}: $load(p, t, x)$;
\item unload a passenger {\it p} from a taxi {\it t} at a specific location {\it x}: $unload(p, t, x)$;
\item move a taxi {\it t} from a location {\it x} to an other {\it y}: $move(t, x, y)$.\\
\end{itemize}

The action, $move(t, x, y)$, can be executed if and only if there is a taxi $t$ at $x$ and $t$ has enough fuel:
\begin{center}
\begin{tabular}[!h]{ll}
\multicolumn{2}{l}{\bf move(t, x, y)} \\
pre & $\{ at(t, x), hasfuel(t, q), (q \geq 10)  \}$ \\
del & $\{ at(t, x), hasfuel(t, q) \}$ \\
add & $\{ at(t, y), hasfuel(t, (q-10)) \}$ \\
\end{tabular}
\end{center}

The action, $load(p, t, x)$, can be executed if and only if there is a taxi $t$ and a passenger $p$ located at the same place $x$:
\begin{center}
\begin{tabular}[!h]{ll}
\multicolumn{2}{l}{\bf load(p, t, x)} \\
pre & $\{ at(p, x), at(t, x) \}$ \\
del & $\{ at(p, x)\}$ \\
add & $\{ in(p, t) \}$ \\
\end{tabular}
\end{center}

The action, $unload(p, t, x)$, can be executed if and only if there is a taxi $t$ containing a passenger $p$ at $x$:
\begin{center}
\begin{tabular}[!h]{ll}
\multicolumn{2}{l}{\bf unload(p, t, x)} \\
pre & $\{ in(p, t), at(t, x) \}$ \\
del & $\{ in(p, t)\}$ \\
add & $\{ at(p, x) \}$ \\
\end{tabular}
\end{center}

\noindent An action $\alpha \in {\cal{A}}$ is described by a transformation operator:
\begin{equation*}
\langle Pre_{\alpha}, Del_{\alpha}, Add_{\alpha} \rangle
\end{equation*}
\begin{itemize}
\item $Pre_{\alpha}$ is the set of predicates that defines the {\it preconditions} of the action $\alpha$;
\item $Del_{\alpha}$ is the set of predicates that defines the knowledge that becomes false after the execution of $\alpha$ (del list);
\item $Add_{\alpha}$ is the set of predicates that defines the knowledge that becomes true after the execution of $\alpha$ (add list).\\
\end{itemize}

\noindent A planning problem is defined by a tuple:
\begin{equation*}
\langle {\cal{E}}, {\cal{O}}, {\cal{G}} \rangle
\end{equation*}
\begin{itemize}
\item $\cal{E}$ defines the knowledge of an agent;
\item ${\cal{O}} = \{ \langle Pre_{\alpha}, Del_{\alpha}, Add_{\alpha} \rangle \ | \ \alpha \in {\cal{A}} \}$ defines  the description of the actions that an agent can execute (i.e., an operators set);
\item $\cal{G}$ defines the goal of an agent, (i.e, a set of predicates).\\
\end{itemize}

\noindent A plan $\pi$ is an ordered list of actions:
\begin{equation*}
\pi = (\alpha_{1}, \ldots, \alpha_{n})
\end{equation*}
where each action $\alpha_{i}$ is an action in $\cal{A}$.\\

Considering a planning problem $\langle {\cal{E}}, {\cal{O}}, {\cal{G}} \rangle$, a plan $\pi = (\alpha_{1}, \ldots, \alpha_{n})$ defines a sequence of $n+1$ {\it world states}
\begin{equation*}
\pi = {\cal{E}}_{0}, {\cal{E}}_{1} \ldots, {\cal{E}}_{n}
\end{equation*}
with
\begin{itemize}
\item ${\cal{E}}_{0} = {\cal{E}}$ and
\item ${\cal{E}}_{i} = ({\cal{E}}_{i-1} - Del_{\alpha_{i}}) \cup Add_{\alpha_{i}} \  (1 \leq i \leq n)$.\\
\end{itemize}

A plan $\pi = (\alpha_{1}, \ldots, \alpha_{n})$ is a solution of a planning problem $\langle {\cal{E}}, {\cal{O}}, {\cal{G}} \rangle$ if and only if:
\begin{enumerate}
\item the preconditions of all actions hold in the previous world state, ${\cal{E}}_{i-1} \models Pre_{\alpha_{i}}$ for $1 \leq i \leq n$;
\item the goal is reached in the final state generated by the plan, ${\cal{E}}_{n} \models {\cal{G}}$.\\
\end{enumerate}

\noindent In our example (paying the taxi is not considered here), a solution plan $\pi$ is
\begin{eqnarray*}
\pi & = & (load(fred, cab38, downtown), \\
& & move(cab38, downtown, park), \\
& & unload(fred, cab38, downtown))
\end{eqnarray*}

\subsection{Assumption-Based Planning model}

The classical planning model presented in the previous section cannot produce plans with assumptions or conjectures. We define a conjecture as a plan that can be executed if some assumptions hold. That leads us to explain the main difference between classical planning and our approach. This difference relies on the action semantics. Indeed, it is not necessary that all preconditions hold in the current state to execute an action. The preconditions that do not hold, build a set of knowledge about the world $\cal{H}$ called {\it assumptions}. \\

\noindent A conjecture $\chi$ is defined as an ordered list of couples
\begin{equation*}
\chi = ( \langle {\cal{H}}_{\alpha_{1}}, \alpha_{1} \rangle, \ldots, \langle {\cal{H}}_{\alpha_{n}}, \alpha_{n} \rangle )
\end{equation*}
where
\begin{itemize}
\item ${\cal{H}}_{\alpha_{i}}$ describes the assumptions that must hold before executing $\alpha_{i}$. If ${\cal{H}}_{\alpha_{i}}$ is an empty set, no assumption is needed to apply $\alpha_{i}$;
\item $\alpha_{i}$ is an action in $\cal{A}$.\\
\end{itemize}

Considering a planning problem $\langle {\cal{E}}, {\cal{O}}, {\cal{G}} \rangle$, a conjecture $\chi$ is an ordered list of $n+1$ world states
\begin{equation*}
\chi = {\cal{E}}_{0}, {\cal{E}}_{1}, \ldots, {\cal{E}}_{n}
\end{equation*}
with
\begin{itemize}
\item ${\cal{E}}_{0} = {\cal{E}} \cup {\cal{H}}_{\alpha_{1}}$ and
\item ${\cal{E}}_{i} = ( \{ {\cal{E}}_{i-1} \cup {\cal{H}}_{\alpha_{i-1}} \} - Del_{\alpha_{i}} ) \cup Add_{\alpha_{i}} \  (1 \leq i \leq n)$.\\
\end{itemize}

In our example, if the taxi has not fuel, a possible conjecture is:
\begin{eqnarray*}
\chi & = & ( \langle \{ \}, load(fred, cab38, downdown) \rangle, \\
& & \langle \{ hasfuel(cab38, 10) \},\\
& & \hspace{0.5cm} move(cab38, downtown, park) \rangle, \\
& & \langle \{ \}, unload(fred, cab38, park) \rangle )
\end{eqnarray*}

To reach its goal, an agent must check all assumptions made by the conjecture. It can count on its teammates competences to make those assumptions become true. In other words, assumptions made by one agent become additional goals to be satisfied. The Assumption-Based Planning is justified by this possible collaborative reasoning in a multi-agent context.

\section{Assumptions generation}
\label{Algorithm}

In classical planning model, operators are applicable if and only if the preconditions of the operators are unifiable with the agent's knowledge base. In order to elaborate conjectures (i.e, plans with assumptions), this constraint is relaxed. We consider that an operator is always applicable even if all preconditions do not hold. Therefore, the application of an operator involves the computation of the lacking facts. This computation is based on the unification algorithm. That is, at least one substitution that makes the preconditions match with some agent's knowledge must be founded. A substitution $\sigma$ is a finite set of the form $\sigma = \{ x_{1} \rightarrow t_{1}, \ldots, x_{n} \rightarrow t_{n}\}$ where every $x_{i}$ is a variable, every $t_{i}$ is a term not equal to $x_{i}$, and $x_{i} \neq x_{j}$ for any $i \neq j$. Let $\sigma$ be a substitution and $p$ be the preconditions of an operator. Then $p\sigma$ is an expression obtained from $p$ by replacing simultaneously each occurrence of the variable $x_{i}$ with the term $t_{i}$.

\subsection{Substitutions computation}

The computation of the set of possible assumptions is described by the algorithm \ref{FindSusbtitutions-algo}. Let $Pre_{\alpha}$ be the preconditions of the action $\alpha$, ${\cal{E}}$ be the agent's knowledge and $\sigma$ an empty substitution at the first step of the algorithm. The {\it FindSubstitutions} procedure computes all possible substitutions from $\cal{E}$ to execute $\alpha$. If $Pre_{\alpha}$ is empty, then $\alpha$ can always be applied and the procedure returns an empty set (line 2). Otherwise, the procedure tries to find recursively the substitution in order to unify $Pre_{\alpha}$ with $\cal{E}$.\\

In this case, the {\it FindSutitutions} procedure unstacks the first precondition $p$ contained in $Pre_{\alpha}$ (line 5) and tries to unify $p$ for each knowledge of the agent (line 7). If the unification process succeeds, there is a substitution $\sigma$ that unifies $p$ with a knowledge of $\cal{E}$. Therefore, the algorithm is recursively launched on $R$, the remaining preconditions of $\alpha$ (line 12).\\

For example, consider the move operator described as follows:
\begin{center}
\begin{tabular}[!h]{ll}
\multicolumn{2}{l}{\bf move(t, x, y)} \\
pre & $\{ at(t, x), hasfuel(t, q), (q \geq 10)  \}$ \\
del & $\{ at(t, x), hasfuel(t, q) \}$ \\
add & $\{ at(t, y), hasfuel(t, (q-10)) \}$ \\
\end{tabular}
\end{center}
The knowledge of the agent $\cal{E}$ is:
\begin{eqnarray*}
& \{ at(cab38, downtown), isloaded(cab38), &\\
& hasfuel(cab38, 10), at(cab74, downtown), &\\
& (not (isloaded(cab74))), hasfuel(cab74, 5), &\\
& at(cab73, downtown) \} &
\end{eqnarray*}
In this case, there are three substitutions:
\begin{itemize}
\item $\sigma_{1} = \{ t \rightarrow cab38, x \rightarrow downtown,  q \rightarrow 10 \}$
\item $\sigma_{2} = \{ t \rightarrow cab74, x \rightarrow downtown,  q \rightarrow 5 \}$
\item $\sigma_{3} = \{ t \rightarrow cab73, x \rightarrow downtown \}$
\end{itemize}

%%%%%%%%%%%%%%%%%%%%%%%%%%%%%%%%%%%%%%%%%%%%%%
% Algorithme pour d�terminer les substitutions
\begin{algorithm}[!htp]

% D�claration des fonctions
\SetKwFunction{FindSubstitutions}{FindSubstitutions}
\SetKwFunction{Unify}{Unify}

% D�claration des variables locales
\SetKwData{Result}{result}
\SetKwData{Failure}{Failure}

% Titre de l'algorithme
\caption{FindSubstitutions($Pre_{\alpha}$, $\cal{E}$, $\sigma$)}
\label{FindSusbtitutions-algo}

% D�but de l'algorithme
\Result $\leftarrow$ an empty set of substitutions \;
\If{$Pre_{\alpha}$ is empty}{
   \Return \Result \;
}
$p \leftarrow$ the first term of $Pre_{\alpha}$\;
$R \leftarrow$ the remaining terms in $Pre_{\alpha}$\;
\ForEach{term $e \in \cal{E}$}{
   $\theta \leftarrow$ \Unify{$p$, $e$, $\sigma$}\;
   \eIf{$\theta$ == \Failure}{
      {\bf continue} \;
   }{
     \ForEach{substitutions $s$ in \FindSubstitutions{$R$, $\cal{E}$, $\sigma \theta$}}{
        add a substitution $s$ in \Result \;
     }
   }
}
\Return \Result\;
\end{algorithm}

\subsection{Assumptions generation}

Each substitution defines a possible executable action in which the parameters are instantiated by the value contained in the substitution. The {\it GenerateAssumption} procedure is given by the algorithm \ref{GenerateAssumptions-algo}. Let $\alpha$ be an action, $\sigma$ a substitution and $\cal{E}$ a set of knowledge. The algorithm applies for each precondition $p$ of $\alpha$ the substitution $\sigma$ and check if $p\sigma$ is contained in the agent's knowledge (line 3). If $p\sigma$ is contained in $\cal{E}$, $p\sigma$ is not an assumption (line 5). In this case, the algorithm has the same behaviour as a classical planner. Otherwise, $p\sigma$ is an assumption needed to execute $\alpha$ (line 8). In the previous example, the application of the three substitutions $\sigma_{1}$, $\sigma_{2}$ and $\sigma_{3}$ on the preconditions of the {\it move} operator produces the following assumptions sets (assumptions are labeled with a~*):\\
\begin{itemize}
\item[A)] $\{$ $at(cab38, downtown),$ $hasfuel(cab38, 10),$ \\
 $isloaded(cab38)$
 $\}$
\item[B)] $\{$ $at(cab74, downtown),$ $hasfuel(cab74, 5),$ \\
 $isloaded(cab74)*,$ $(q \geq 10)$ $\}$
\item[C)] $\{$ $at(cab73, downtown),$ $hasfuel(cab73, q)*,$ \\
 $isloaded(cab73)*,$ $(q \geq 10)*$ $\}$ \\
\end{itemize}

The result A of the substitution $\sigma{1}$ application is a subset of the agent's knowledge. Therefore, no assumption is needed to execute the {\it move} action. However, the results B and C of the $\sigma_{2}$ and $\sigma_{3}$ applications contain assumptions. The {\it GenerateAssumptions} procedure distinguishes three kinds of assumptions:
\begin{enumerate}
\item {\bf Hypothesis generation:} the substitution can generate literals that do not belong to the current state, e.g. in C)  {\it isloaded(cab73)}. Then, those literals are added to the current state as {\it hypothesis}. This means that expressions missing from the current state are not considered as false but rather as unknown. Hypothesis can also contain variable symbols not instantiated if no instance can be found by the substitution function (e.g., {\it ?q} in C)  {\it hasfuel(cab73, ?q)}). This means that {\it cab73} has an unknown quantity of fuel, but this unknown does not prevent the planning process to proceed (e.g. assessment of {\it ?q} remains an open issue to be fixed by another agent);\\
\item {\bf Fact negation:} if an atom in the substitution is the negation of a fact in the current state, then this fact is withdrawn and replaced by its negation (e.g., the ground atom in B) {\it isloaded(cab74)}). In that case, the agent knows that it is reasoning by contradiction and that it bets on its teammates ability to change the world consistently (in that example, by loading {\it cab74});\\
\item {\bf Constraint violation:} as in fact negation, constraints can be violated e.g. in B) {\it (?q $>=$ 10.00)}.
\end{enumerate}

%%%%%%%%%%%%%%%%%%%%%%%%%%%%%%%%%%%%%%%%%%%%%%
% Algorithme pour g�n�rer les hypoth�ses
\begin{algorithm}[!htp]

% D�claration des fonctions
\SetKwFunction{FindSubstitutions}{FindSubstitutions}

% D�claration des variables locales
\SetKwData{Failure}{Failure}
\SetKwData{Result}{result}

% Titre de l'algorithme
\caption{GenerateAssumptions($\alpha$, $\sigma$, $\cal{E}$)}
\label{GenerateAssumptions-algo}

% D�but de l'algorithme
\Result $\leftarrow$ an empty set of assumptions \;
$Pre_{\alpha} \leftarrow$ the preconditions of $\alpha$ \;
\ForEach{precondition $p \in Pre_{\alpha}$}{

   {\small\bfseries\itshape\ttfamily // $p \sigma$ is an agent's knowledge} \;

   \lIf{$p \sigma \in {\cal{E}}$}{
      {\bf continue} \;
   }

   \BlankLine

   {\small\bfseries\itshape\ttfamily // $p \sigma$ is an assumption} \;

   \Else{
      add $p \sigma$ in \Result \;
   }
}
\Return \Result \;
\end{algorithm}

\section{Planning algorithm}
\label{planning-algo}

The Assumption-Based Planning algorithm principle relies on a domain independent planning mechanism, {\sc Htn} (Hierarchical Transition Network). In HTN planner \cite{nau-03}, the objective is not to achieve a set of goals but to perform some sets of {\it tasks}. The agent's input includes a set of operators similar to those used in classical planning \cite{finkes-71} and also a set of {\it methods}, each of which is a prescription on how to decompose some tasks into some sets of subtasks.  The agent proceeds by decomposing {\it non-primitive tasks} recursively into smaller and smaller subtasks, until {\it primitive tasks}, that can be performed directly by planning operators, are reached.

\subsection{Primary notions}

A {\it primitive action} $\alpha$ is described by an operator $\langle name_{\alpha}, Pre_{\alpha}, Del_{\alpha}, Add_{\alpha} \rangle$ where $name_{\alpha}$ is the name of the primitive action, $Pre_{\alpha}$ is the preconditions set needed to execute $\alpha$, $Del_{\alpha}$ and $Add_{\alpha}$ define respectively the set of effects to delete and to add to the agent's knowledge. An operator is executed when all preconditions $Pre_{\alpha}$ are satisfied in the current knowledge state of the agent. The operator execution involves the modification of the current state according to the effects contained in $Del_{\alpha}$ et $Add_{\alpha}$. An operator can be applied if there is a substitution $\sigma$ for $Pre_{\alpha}$ such as $Pre_{\alpha}\sigma$ is instantiated. $n_{\alpha} \sigma$ defines the action that can be executed for each variable replaced by a value contained in $\sigma$. For example, the operator $load(p, t, x)$:
\begin{center}
\begin{tabular}[!h]{ll}
\multicolumn{2}{l}{\bf load(p, t, x)} \\
pre & $\{ at(p, x), at(t, x) \}$ \\
del & $\{ at(p, x)\}$ \\
add & $\{ in(p, t) \}$ \\
\end{tabular}
\end{center}
and the agent's knowledge:
\begin{equation*}
\{ \ at(cab38, downtown), at(fred, donwtown) \ \}
\end{equation*}
There is a substitution $\sigma$ that binds the variables $p$, $t$ et $x$ to the constant values $fred$, $cab38$ and $downtown$. Therefore, the action {\it load}$(fred,$ $cab38, downtown)$ can be executed.\\

A {\it compound action} $\alpha$ is described by a method $\langle n_{\alpha}, Pre_{\alpha}, Act_{\alpha} \rangle$ where $n_{\alpha}$ is the name of $\alpha$, $Pre_{\alpha}$ is the preconditions set needed to apply $\alpha$, $Act_{\alpha}$ defines a list of actions to execute (i.e., the method body). A method can be executed when all preconditions $Pre_{\alpha}$ are satisfied in the current state. A method execution involves the execution of all actions contained in the ordered actions list $Act_{\alpha}$. A method can be applied if there is a substitution $\sigma$ for $Pre_{\alpha}$ with $Pre_{\alpha} \sigma$ instantiated. $Act_{\alpha} \sigma$ defines the list of actions to execute the method. For example, let a method {\it move-passenger}$(p, x, y)$ that move a passenger $p$ from a location $x$ to $y$:
\begin{center}
\begin{tabular}[!h]{ll}
\multicolumn{2}{l}{\bf move-passenger(p, x, y)} \\
pre & $\{ at(p, x), at(t, x) \}$ \\
act & $\{ load(p, t,  x), move(t, x, y), unload(p, t, y) \}$ \\
\end{tabular}
\end{center}
and the agent's knowledge:
\begin{equation*}
\{ \ at(cab38, downtown), at(fred, downtown) \ \}
\end{equation*}
There is a substitution $\sigma$ that binds each variables $p$, $x$, $y$ and $t$ with the constant values $fred$, $downtown$, $park$ and $cab38$. Therefore, the action {\it move-passenger}$(fred, downtown, park)$ can be executed. This execution involves the execution of three actions contained in the body of the operator:
\begin{enumerate}
\item $load(fred, cab38, downtown)$;
\item $move(cab38, downtown, park)$;
\item $unload(fred, cab38, park)$.\\
\end{enumerate}

A conjecture is an ordered list of primitive actions. If $\chi$ is a conjecture and $\cal{E}$ a knowledge state,  $\chi({\cal{E}})$ is the state reached after the execution of $\chi$ from $\cal{E}$.\\

A planning problem is defined by $\langle {\cal{E}}, {\cal{O}}, {\cal{G}} \rangle$:
\begin{itemize}
\item $\cal{E}$ defines the agent's knowledge. This knowledge is described by a set of propositions;
\item $\cal{O}$ defines the set of operators or methods;
\item $\cal{G}$ defines the ordered list of goals that must be reached by the agent.\\
\end{itemize}

The set of solution conjectures ${\cal{C}}({\cal{E}}, {\cal{O}}, {\cal{G}})$ of a planning problem can be recursively defined:
\begin{itemize}
\item If $\cal{G}$ is an empty set, the empty conjecture is returned.
\item Otherwise, let $\alpha$ be the first task or goal of $\cal{G}$, and $\cal{R}$ be the remaining goals:
\begin{enumerate}
\item If $\alpha$ is a primitive action and there is a conjecture $\chi_{1}$ to reach $\alpha$ then,
\begin{equation*}
{\cal{C}}({\cal{E}}, {\cal{O}}, {\cal{G}}) = \{append(\chi_{1}, \chi_{2}) \ | \ \chi_{2} \in {\cal{C}}({\cal{E}}, {\cal{O}}, {\cal{R}})\}
\end{equation*}
\item If $\alpha$ is a primitive action and there is not a conjecture $\chi$ to reach $\alpha$ then,
\begin{equation*}
{\cal{C}}({\cal{E}}, {\cal{O}}, {\cal{G}}) = \emptyset
\end{equation*}
\item If $\alpha$ is a composed action then,
\begin{equation*}
{\cal{C}}({\cal{E}}, {\cal{O}}, {\cal{G}}) = {\cal{C}}({\cal{E}}, {\cal{O}}, append(Act_{\alpha}, {\cal{R}})
\end{equation*}
where $Act_{\alpha}$ defines the actions list to be executed in order to realise $\alpha$.
\end{enumerate}
\end{itemize}

\subsection{Principle}

Until now, we have considered that a method or an operator was always executable, even if all their preconditions were not completely held and we have presented the algorithm to compute assumptions. The number of possible assumptions is potentially unlimited. Therefore, it is not possible to compute all possible conjectures. The purpose of the Assumption-Based Planning algorithm is to find conjectures that make the fewest assumptions. To that end, the Assumption-Based Planning algorithm is based on a reachable states search space. This states space is stored in a tree called the {\it conjecture tree}. \\

This tree contains the different steps of the agent's reasoning. Each node stands for a state that can be reached after the execution of an action $\alpha_{i}$ in a conjecture $\chi = (\alpha_{1}, \ldots, \alpha_{n})$ A node $n_{i}$ is defined by $\langle {\cal{E}}_{i}, {\cal{A}}_{i}, w_{i} \rangle$:
\begin{itemize}
\item ${\cal{E}}_{i}$ is a world state (i.e., a set of propositions fully or partially instantiated);
\item ${\cal{A}}_{i}$ is a list of remaining actions to execute at this step of the reasoning;
\item $w_{i}$ is the valuation of the node (i.e., the number of assumptions given from the root node of the conjecture tree to this node).\\
\end{itemize}

The edges define the possible transitions between the different world states. A transition is labeled by a method or operator name and the possible assumptions needed to reach it.\\

The general procedure to produce a conjecture is given by the algorithm \ref{FindConjecture-algo}. The {\it FindConjecture} procedure takes as parameters a planning problem: $\cal{E}$ the initial state (i.e., the agent's knowledge), $\cal{O}$ a list of operators and $\cal{G}$ the ordered list of goals. The root node of the conjecture tree is defined by the node $\langle {\cal{E}}, {\cal{G}}, 0 \rangle$,  ${\cal{E}}$ is the initial set of the agent's knowledge,  ${\cal{G}}$ the list of goals and null valuation.\\

The algorithm can be split in two different steps: the conjecture tree expansion (line 3) which represents the reachable states space and the conjecture extraction (line 4).

%%%%%%%%%%%%%%%%%%%%%%%%%%%%%%%%%%%%%%%%%%%%%%
% Algorithme pour calculer une conjecture
\begin{algorithm}[!htp]

% D�claration des fonctions
\SetKwFunction{ExpandConjectureTree}{ExpandConjectureTree}
\SetKwFunction{ExtractConjecture}{ExtractConjecture}

% D�claration des variables locales
\SetKwData{Failure}{Failure}
\SetKwData{CTree}{CTree}
\SetKwData{Root}{root}

% Titre de l'algorithme
\caption{FindConjecture($\cal{E}$, $\cal{O}$, $\cal{G}$)}
\label{FindConjecture-algo}
\BlankLine

\Root $\leftarrow$ $\langle {\cal{E}}, {\cal{G}}, 0 \rangle$ \;
\CTree $\leftarrow$ create a conjecture tree with root node \Root \;
\ExpandConjectureTree{\CTree, \Root} \;
\Return \ExtractConjecture{\CTree} \;

\BlankLine
\end{algorithm}

\subsection{Conjecture tree expansion}

%%%%%%%%%%%%%%%%%%%%%%%%%%%%%%%%%%%%%%%%%%%%%%
% Algorithme de construction de l'arbre de conjectures
\begin{algorithm}[!h]

% D�claration des fonctions
\SetKwFunction{ExpandConjectureTree}{ExpandConjectureTree}
\SetKwFunction{FindSubstitutions}{FindSubstitutions}
\SetKwFunction{GenerateAssumptions}{GenerateAssumptions}
\SetKwFunction{Append}{Append}

% D�claration des variables locales
\SetKwData{Failure}{Failure}
\SetKwData{BestNode}{bestNode}

% Titre de l'algorithme
\caption{ExpandConjectureTree($CTree$, $n_{i}$)}
\label{ExpandConjectureTree-algo}

% D�but de l'algorithme
\BlankLine

{\small\bfseries\itshape\ttfamily // Extract the first action $\alpha$ contained in}
{\small\bfseries\itshape\ttfamily // the current node $n_{i}$}

$\alpha \leftarrow$ the first action contained in $n_{i}$ \;
${\cal{R}} \leftarrow$ the remaining actions contained in $n_{i}$ \;

\BlankLine

{\small\bfseries\itshape\ttfamily // Compute the possible substitutions from} \\
{\small\bfseries\itshape\ttfamily // $Pre_{\alpha}$, $\alpha$ preconditions, and ${\cal{E}}_{i}$, the world}\\
{\small\bfseries\itshape\ttfamily // of the current node $n_{i}$ and $\sigma$ an empty}\\
{\small\bfseries\itshape\ttfamily // substitution}

$\Sigma \leftarrow$ \FindSubstitutions{$Pre_{\alpha}$, ${\cal{E}}_{i}$, $\sigma$} \;

\BlankLine

{\small\bfseries\itshape\ttfamily // Create a new node $n_{i+1}$ for each}\\
{\small\bfseries\itshape\ttfamily // each applicable action}

\ForEach{substitution $\sigma \in \Sigma$}{

{\small\bfseries\itshape\ttfamily // Compute the assumptions needed to}\\
{\small\bfseries\itshape\ttfamily // apply $\alpha$}

${\cal{H}}_{\alpha} \leftarrow$ \GenerateAssumptions{$\alpha$, $\sigma$, ${\cal{E}}_{i}$} \;

\BlankLine

{\small\bfseries\itshape\ttfamily // Check if the assumption generated}\\
{\small\bfseries\itshape\ttfamily // are legal}

\lIf{$\exists \ h \in {\cal{H}}_{\alpha}$ with $h$ a legal assumption}{{\bf continue}} \;

\BlankLine

{\small\bfseries\itshape\ttfamily // $\alpha$ is a primitive action}

\If{$\alpha$ is primitive}{
${\cal{E}}_{i+1} \leftarrow ( \{ {\cal{E}}_{i} \cup {\cal{H}_{\alpha}} \} - Del_{\alpha} ) \cup Add_{\alpha}$ \;
${\cal{A}}_{i+1} \leftarrow {\cal{R}}$ \;
$w_{i+1} \leftarrow w_{i} + |{\cal{H}}_{\alpha}|$ avec $|{\cal{H}}_{\alpha}|$ the number of elements in  ${\cal{H}}_{\alpha}$ \;
}

\BlankLine

{\small\bfseries\itshape\ttfamily // $\alpha$ is a composed action}

\Else {
${\cal{E}}_{i+1} \leftarrow {\cal{E}}_{i} \cup {\cal{H}_{\alpha}} $ \;
${\cal{A}}_{i+1} \leftarrow$ \Append{$Act_{\alpha}$, ${\cal{R}}$} \;
$w_{i+1} \leftarrow w_{i} + |{\cal{H}}_{\alpha}|$ with $|{\cal{H}}_{\alpha}|$ the number of elements in ${\cal{H}}_{\alpha}$ \;
}

\BlankLine

{\small\bfseries\itshape\ttfamily // Add a new node $n_{i+1}$ to the} \\
{\small\bfseries\itshape\ttfamily // conjecture tree}

add a new node $n_{i+1} = \langle {\cal{E}}_{i+1}, {\cal{A}}_{i+1}, w_{i+1} \rangle$ \;
}
\BlankLine

{\small\bfseries\itshape\ttfamily // Compute the node with few assumptions} \\
{\small\bfseries\itshape\ttfamily // and run recursively the procedure} \\
{\small\bfseries\itshape\ttfamily // \ExpandConjectureTree}

\BestNode $\leftarrow$ the node with few assumptions in $CTree$ \;
\ExpandConjectureTree{\BestNode} \;

\BlankLine
\end{algorithm}

Unlike {\sc Htn} recursive algorithm, the conjecture tree expansion is not a simple depth first exploration: the computation of the conjecture with the fewest assumptions is equivalent to a minimization problem. In order to solve this problem, the expansion algorithm is based on the ``branch and bound'' algorithm. The nodes stored in the conjecture tree  are valuated by the number of assumptions made (see algorithm \ref{ExpandConjectureTree-algo}) to reach it and the node with the weakest valuation is recursively chosen at the expansion step to expand the conjecture tree until a leaf is found.\\

The expansion procedure is described by the algorithm \ref{ExpandConjectureTree-algo}. From the current exploration node $n_{i} = \langle {\cal{E}}_{i}, {\cal{A}}_{i}, w_{i} \rangle$, the {\it ExpandConjecture} procedure tries to apply the first action $\alpha$ contained in ${\cal{A}}_{i}$ (line 1). Then it computes the possible substitutions $\Sigma$ for the preconditions of $\alpha$ according to the knowledge ${\cal{E}}_{i}$ (line 6). For each substitution, the procedure generates the assumptions  ${\cal{H}}_{\alpha}$ needed to apply $\alpha$ (line 10). If $\alpha$ is a primitive action (line 14), then it adds a child node $n_{i+1}$ to the current node such as:
\begin{equation*}
n_{i+1} = \langle \ (\{ {\cal{E}}_{i} \cup {\cal{H}_{\alpha}} \} - Del_{\alpha} ) \cup Add_{\alpha}, {\cal{R}}, w_{i} + |{\cal{H}}_{\alpha}| \ \rangle
\end{equation*}
where $|{\cal{H}}_{\alpha}|$ defines the number of assumptions needed to apply $\alpha$. Otherwise (line 20), $\alpha$ is a composed action and the procedure adds a child node to the current node such as:
\begin{equation*}
n_{i+1} = \langle \ \{ {\cal{E}} \cup {\cal{H}_{\alpha}} \}, append(Act_{\alpha}, {\cal{R}}), w_{i} + |{\cal{H}}_{\alpha}| \ \rangle
\end{equation*}

Finally, the procedure evaluates the new current node by computing the node with the lowest valuation in the conjecture tree (line 30) and runs recursively the expansion procedure with the new node (line 31).\\

When assumptions are generated by the algorithm, the expansion procedure checks if the assumptions are legal (line 11). An assumption is legal if its assumption predicates was described in the planning domain as hypothetical. The legality notion was added to our algorithm to specify states of the world that can be hypothetical. This mechanism allows to reduce the states space to explore, limiting the size of the conjecture tree.

\subsection{Conjecture Extraction}

The second step is the conjecture extraction from the conjecture tree (see Algo. \ref{ExtractConjecture-algo}). A conjecture is represented by a branch (i.e. a path from the root node to a leaf). First, the {\it ExtractConjecture} procedure computes among the set of solution nodes (i.e., the leaf of the conjecture tree) the node that makes the fewest assumptions (line 2). For each edge of the branch from the chosen solution node to the root node (line 3), the procedure checks the type of transition (i.e., primitive or composed action). If the edge is labeled with a primitive one, then the action and their assumptions are added to the conjecture (line 5). Otherwise, only the assumptions are added to the conjecture (line 7).

%%%%%%%%%%%%%%%%%%%%%%%%%%%%%%%%%%%%%%%%%%%%%%
% Algorithme d'extraction de la conjecture
\begin{algorithm}[!htp]

% D�claration des fonctions

% D�claration des variables locales
\SetKwData{Failure}{Failure}
\SetKwData{CTree}{CTree}

% Titre de l'algorithme
\caption{ExtractConjecture($CTree$)}
\label{ExtractConjecture-algo}

$\chi \leftarrow$ an empty conjecture  \;
$n \leftarrow$ the leaf solution node with the fewest assumptions \;

\While{$n$ is not the root node of \CTree} {
   \eIf{$n$ was built with a primitive action $\alpha_{i}$}{
       add action $\langle {\cal{H}}_{\alpha_{i}}, \alpha_{i} \rangle$ to $\chi$\;
    }{
       add the assumptions ${\cal{H}}_{\alpha_{i}}$ to the action $\alpha_{i-1}$ contains in $\chi$ \;
    }
   $n \leftarrow$ the father node of $n$ \;
}
\Return $\chi$ \;
\BlankLine
\end{algorithm}

\section{Discussion}
\label{discussion}

{\bf Soundness and completeness:} The algorithm tries to decompose the initial goal in an ordered list of primitive tasks. As in {\sc Shop}, for a finite search space, the construction of the conjecture tree is sound and complete. For infinite search spaces, it is also complete due to the iterative-deepening conjecture tree construction. However, our algorithm is more greedy because, when assumptions must be done, more nodes are created. The primary tests on our JAVA implementation highlight results in the same order of magnitude than the {\sc JShop} algorithm (the JAVA implementation of {\sc Shop} algorithm).\\

{\bf Search limitation:} The number of allowed assumptions can be bounded in order to end the search process at an arbitrary limit. When the limit is set to 0, the algorithm is equivalent to {\sc Shop}. This can be used to adapt our algorithm to the system capabilities and find conjecture with more and more assumptions.\\

{\bf Choice of a planning system:} There are many different planning systems (e.g., planning based on Binary Decision Diagrams \cite{jensen-00}, Mutex \cite{blum-97}, heuristic search \cite{do-90}, constraints satisfaction \cite{kautz-99} and so forth). However, {\sc Shop} is well-suited for assumptions generation thank to the substitution procedure that allows to compare the agent's knowledge with the preconditions necessary to trigger an operator or a method. We found out that assumptions generation turned out to be much more difficult in planners like {\sc GraphPlan} \cite{blum-97}.\\

{\bf Adequacy with interactions:} In teamwork context, a conjecture can be refuted by another agent; if no repair is found, the conjecture must be abandoned. In this case, the cost of providing another conjecture is low because agents can rely on the conjecture tree already computed and resume their exploration.

\section{Related work}
\label{related-works}

The problem of constructing plans in a distributed environment has been approached from two different directions. One approach makes an emphasis on the problem of controlling and coordinating the actions of multiple agents in a shared environment. The others approaches focus on planning and how it can be extended into distributed environment, where the process of formulating a plan could involve actions and interactions of many participants. The planning approach is the nearest to Assumption-Based Planning.\\

The first approach objective is not to form a good collective plan, but rather to ensure that the agent's local objective will be met by this plans. This approach based on {\sc Bdi} models formalizes the distributed planning process using the mental states of the agents \cite{bratman-87,pollack-90,rao-91}. These approaches emphasize the necessity for a group of agents to share a {\it joint intention} in order to reach a goal \cite{cohen-90,levesque-90}. They have been validated in projects such as {\sc Steam} \cite{tambe-97}. Another cognitive approach \cite{grosz-96} formalizes the coordination process by two kinds of intentions: the agent's {\it intention to} do an action or a plan and the agent's {\it intention that} some propositions hold. Then, shared plans are generated by combining predefined plans or ``recipes''.\\

The second approach \cite{corkill-79,lesser-81} places the problem of forming a plan as the ultimate objective and is typically carried out by agents that have been endowed with shared goals and representation. This approach can be divided in three distinct steps: the decomposition step of tasks into subtasks \cite{durfee-87,lansky-95}, the allocation step \cite{smith-88} and the execution and conflict resolution step. This formal division is sometimes difficult to implement because the three steps are not independent in many cases: conflict resolution can imply to re-allocate several subtasks or to seek another decomposition and, therefore, deadlocks must be carefully dispelled. Several frameworks focus on the detection of relations between plans \cite{martial-90}. Two kinds of relations are identified: positive relations (e.g., redundant tasks) or negative relations (e.g., resources conflicts). Conflicts are solved in many ways: by negotiation \cite{zlotkin-90}, by argumentation \cite{kraus-93,tambe-99}, by synchronization \cite{clement-99,clement-03} etc. More recently, the systems {\sc Dsipe} \cite{desjardins-00} that is a distributed version of SIPE-2 or the dMARS project \cite{inverno-04} based on a PRS architecture, are interested in solving real-world planning problems and, to that end, argue for the use of domain knowledge in planning. \\

Although these coordination mechanisms bring us a number of answers to make a group of agents work together as a team, they show limitations to solve cooperative task when the goals of the agents are not well proportioned. These limitations can be explained by the fact that coordination mechanisms are not interleaved in the planning process and often use predating plans or recipes libraries. Another limitation is due to the difficulty to take into account the necessary uncertainty in real-world  because of the increasing planning complexity coming from incomplete information.

\section{Conclusion}

The Assumption-Based Planning model outlined in this paper relies on plan production and revision by conjecture/refutation cycles. We presented one of the most important elements of this project: the Assumption-Based Planner. This process is based on a {\sc Htn} planner, {\sc Shop}. For a given initial task, the algorithm seeks the less hypothetical plan (i.e, conjecture). It breaks down the initial task recursively into simple operators. At each step, if no complete substitution is found, the operators or methods constraints are relaxed in order to compute the assumptions necessary to push the plan elaboration further. The construction of the conjecture tree is guided by the valuation of each node in terms of assumptions guaranteeing to find the most reasonable conjecture. As far as teamwork is concerned, each time a conjecture is definitively abandoned, a new one can be proposed from the ongoing conjecture tree. \\

Thus, Assumption-Based Planning model merges in the collaborative plan generation, the decomposition and the coordination steps. Moreover, the Assumption-Based Planning includes in the agents' reasoning the notion of uncertainty and allows to compose the agents competences. The argumentation is used to structure the multi-agent reasoning as a collaborative investigation process and not as a negotiation one. From our point of view, this approach is suitable for applications in which agents share a common goal and in which the splitting of the planning and the coordination steps (when agents have independent goals, they locally generate plans and then solve their conflicts) becomes difficult due to the agents strong interdependence. Our target applications are the composition of web services, the cooperation of video games characters and the dynamic reconfiguration of {\sc Gui} components. \\

Our future research direction will be the formalization of the conjecture/refutation protocol: the corresponding dialog games have to define interaction rules ensuring the convergence of the multi-agent planning process towards a mutually accepted conjecture; and the identification of different kinds of domain-independent plan refutation to build robust plans.

\end{document}